\documentclass{article}

\usepackage{arxiv}

\usepackage[utf8]{inputenc} 
\usepackage[T1]{fontenc}    
\usepackage{hyperref}       
\usepackage{url}            
\usepackage{booktabs}       
\usepackage{amsfonts}       
\usepackage{nicefrac}       
\usepackage{microtype}      
\usepackage{lipsum}		
\usepackage{graphicx}
\usepackage{natbib}
\usepackage{doi}
\usepackage{multirow}
\usepackage[acronym, shortcuts]{glossaries}
\usepackage{multirow}
\usepackage[utf8]{inputenc} 
\usepackage[T1]{fontenc}    
\usepackage{url}            
\usepackage{booktabs}       
\usepackage{amsfonts}       
\usepackage{nicefrac}       
\usepackage{microtype}      
\usepackage[acronym, shortcuts]{glossaries}
\usepackage{graphicx}
\usepackage{caption}
\usepackage{subcaption}
\usepackage{amsmath}
\usepackage{tikz}
\usepackage{amssymb}

\newacronym{fao}{FAO}{Food and Agriculture Organization}
\newacronym{td}{TD}{Temporal Dropout}
\newacronym{eo}{EO}{Earth Observation}
\newacronym{s2}{S2}{Sentinel-2}
\newacronym{rs}{RS}{Remote Sensing}
\newacronym{lstm}{LSTM}{Long-Short Term Memory}
\newacronym{rnn}{RNN}{Recurrent Neural Network}
\newacronym{cnn}{CNN}{Convolutional Neural Network}
\newacronym{mlp}{MLP}{Multi-layer Perceptron}
\newacronym{ml}{ML}{Machine Learning}
\newacronym{et}{ET}{Evapotranspiration}

\title{Exploring Physics-Informed Neural Networks for Crop Yield Loss Forecasting}

\author{%
  Miro Miranda$^{1,2,*}$, Marcela Charfuelan$^{2}$, and Andreas Dengel$^{1,2}$ \\ 
  $^{1}$Department of Computer Science, University of Kaiserslautern-Landau,  Kaiserslautern, Germany \\
  $^{2}$German Research Center for Artificial Intelligence, Kaiserslautern, Germany \\
    $^{*}$\texttt{miro.miranda\_lorenz}\texttt{@dfki.de} \\
}

\renewcommand{\shorttitle}{Tackling Climate Change with Machine Learning: workshop at NeurIPS 2024}

\hypersetup{
pdftitle={A template for the arxiv style},
pdfsubject={q-bio.NC, q-bio.QM},
pdfauthor={David S.~Hippocampus, Elias D.~Striatum},
pdfkeywords={First keyword, Second keyword, More},
}

\begin{document}
\maketitle

\begin{abstract}
In response to climate change, assessing crop productivity under extreme weather conditions is essential to enhance food security. Crop simulation models, which align with physical processes, offer explainability but often perform poorly. Conversely, machine learning (ML) models for crop modeling are powerful and scalable yet operate as black boxes and lack adherence to crop growth’s physical principles. To bridge this gap, we propose a novel method that combines the strengths of both approaches by estimating the water use and the crop sensitivity to water scarcity at the pixel level. This approach enables yield loss estimation grounded in physical principles by sequentially solving the equation for crop yield response to water scarcity, using an enhanced loss function. Leveraging Sentinel-2 satellite imagery, climate data, simulated water use data, and pixel-level yield data, our model demonstrates high accuracy, achieving an $R^2$ of up to 0.77—matching or surpassing state-of-the-art models like RNNs and Transformers. Additionally, it provides interpretable and physical consistent outputs, supporting industry, policymakers, and farmers in adapting to extreme weather conditions.\end{abstract}


\section{Introduction}
\begin{figure*}[!ht]
  \centering
\begin{subfigure}{.55\textwidth}
  \centering
  \includegraphics[width=\linewidth]{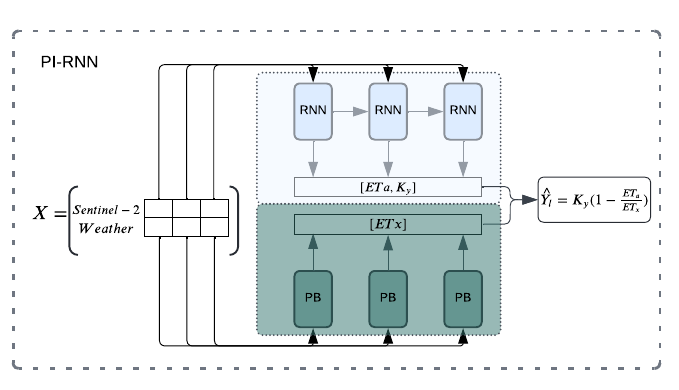}
  \caption{Physics-Informed Model}
  \label{fig:early_fusion_model}
\end{subfigure}%
\begin{subfigure}{.45\textwidth}
  \centering
  \includegraphics[width=\linewidth]{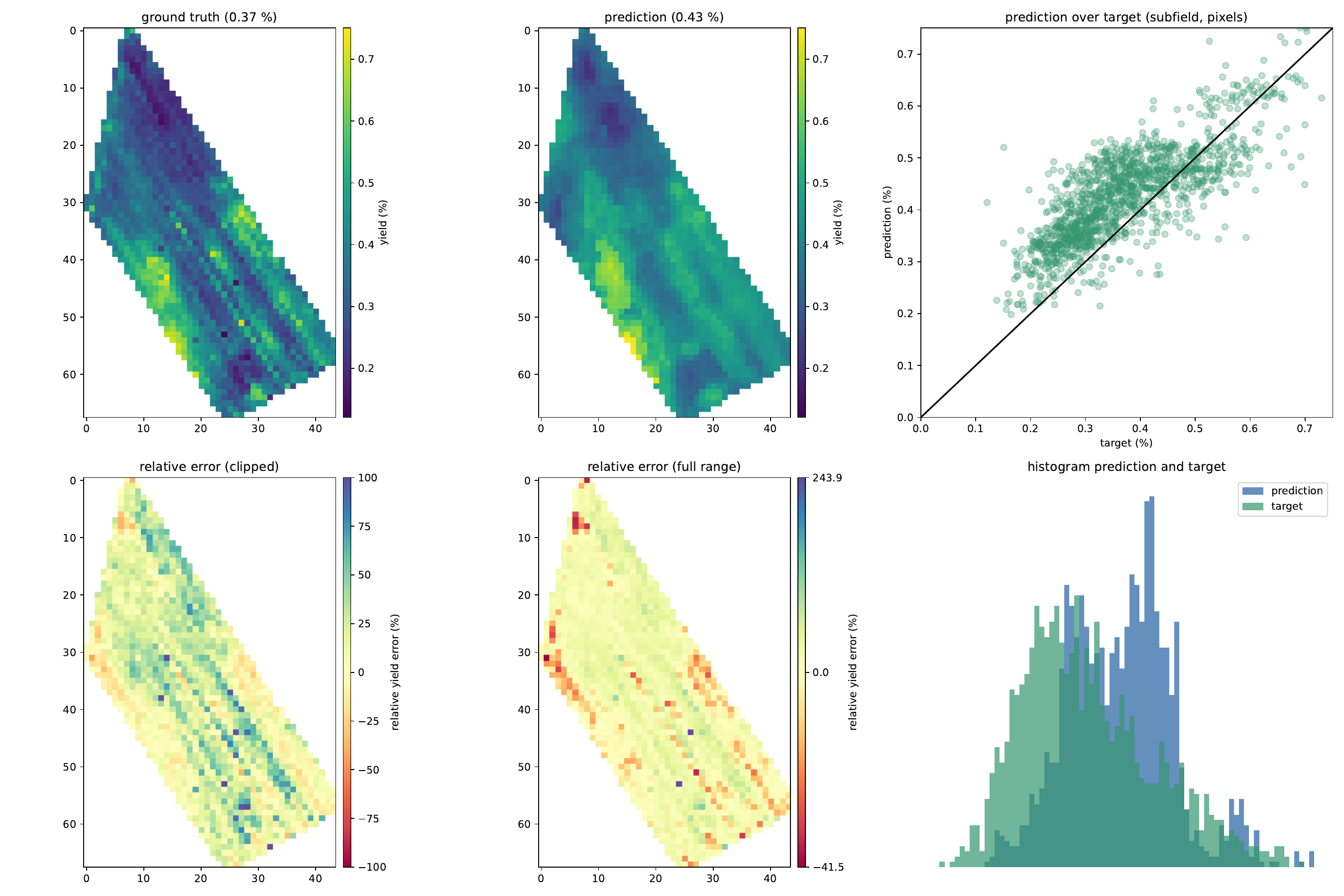}
  \caption{Prediction, target and yield loss distribution.}
  \label{fig:field_evaluation}

\end{subfigure}
\caption{(a) Framework for physics-informed yield loss forecasting. Data modalities, including Sentinel-2 imagery and climate variables, are used to train an RNN that predicts biophysical properties for each time point (actual water use ($ET_a$), and the crop susceptibility to water scarcity ($K_y$)). A simulation model predicts  the maximum water use ($ET_x$). The predicted biophysical properties are used to calculate the actual yield loss ($Y_l$). We leverage prior knowledge describing the relationship between water use and the relative yield loss. (b) Performance plots for visual inspection of a single field. Yield data from cereals in Switzerland is shown, harvested in 2020.}
\label{fig:main_fig}
\end{figure*}
In response to climate change, closing the gap between potential and actual yields is an urgent task to sustain food security \citep{fischer2012crop}. Extreme weather conditions like droughts and floodings are pressing challenges in the agricultural sector, directly affecting productivity, and causing yield losses \citep{arora2019impact, molotoks2021impacts}. Digital agriculture has emerged as a key strategy to address these challenges, providing tools for timely, informed decision-making by policymakers, industry stakeholders, and farmers \citep{roscher2023data}. 
The response of crop yields to water availability has been a central focus of research for decades, serving as a critical parameter in assessing crop resilience under extreme climate conditions \citep{pereira2015crop}. Traditionally, simulation models, also known as process-based models, have been employed to capture this relationship. These models build upon biological and physical principles to offer high explainability, supporting decisions in areas such as irrigation, fertilization, pest management, and disease control. However, crop simulation models often struggle with high-resolution, multidimensional data, are computationally intensive, and require calibration, rendering their applicability to high spatial resolution  (pixel-level) unfeasible. Furthermore, they are often simplified representations of reality, relying on approximations to maintain computational efficiency \citep{kang2009climate}, oftentimes resulting in inaccurate performances \citep{leng2020predicting}. Therefore, machine learning (ML) models are increasingly utilized for crop productivity estimation \citep{van2020crop}. Recent studies demonstrate impressive scalability and accuracy, even at pixel resolution \citep{helber2023crop, pathak}, by handling complex and multidimensional data \citep{miranda2024multi}. However, ML models are often criticized for their black-box characteristics, limiting their interpretability \citep{castelvecchi2016can}. Additionally, ML are seldomly designed to follow underlying physical principles of plant growth \citep{he2023physics}, thereby hardly ever providing meaningful intermediate outputs, which are essential for tracking crop responses to changing environmental conditions over time. There is a growing demand to integrate the strengths of data-driven ML approaches with the interpretability of simulation models \citep{dash2022review, kang2009climate}. While some recent studies attempt to combine ML and prior knowledge for yield prediction, also under water-limited conditions \citep{shuai2022subfield}, these approaches generally focus on data enrichment rather than on producing interpretable and physical consistent sequential outputs of crop physiology, similar to simulation models. Consequently, they often retain the black-box limitations of conventional ML models, providing limited insight into crop physiology by potentially violating governing physical principles. \\
This study seeks to bridge the gap between interpretable simulation models and high-performance ML models by introducing a physics-informed approach to crop yield loss forecasting. We argue that we can learn more precisely the water use and the crop susceptibility to water scarcity, by building upon Sentinel-2 (S2) multispectral imagery, climate data, and simulations, coupled with pixel-level yield data from cereal crops harvested between 2017–2021 in Switzerland. We demonstrate yield loss estimation at the pixel level, grounded on physical principles by sequentially solving the equation for crop yield response to water scarcity \citep{doorenbos1979yield}, using an enhanced loss function. Initial results indicate high potential, achieving an $R^2$ of up to 0.77, matching or surpassing the performance of state-of-the-art (SOTA) ML models for yield prediction, including RNNs and Transformers, while providing interpretable and physical consistent outputs, supporting industry, policymakers, and farmers in adapting to extreme weather conditions.
\section{Methodology}
\paragraph*{Problem Definition} The water use, also called \textit{evapotranspiration} (ET), is the sum of all biophysical processes in which liquid water is converted to water vapor from various surfaces, including topsoil and vegetation. For a specific crop type, the Food and Agriculture Organization (FAO) differentiates between the maximum ($ET_x$) and the actual ET ($ET_a$). The former represents the water use under standard and non-limiting environmental conditions and is solely impacted by climate conditions and crop specifics, achieving full productivity, i.e., disease-free, well-fertilized, and under optimum soil water conditions. 
In contrast, the $ET_a$ represents the water use under limiting environmental conditions, resulting in a reduction in water use and ultimately causing a reduction of crop productivity. This reduction in productivity is defined as yield loss and typically expressed in relative terms. Various factors cause productivity limiting conditions, including soil infertility, soil salinity, limited soil water content, diseases, and poor management. Especially, in the light of extreme weather conditions, the frequency of severe droughts and floodings is expected to increase, causing either water scarcity or water abundance, negatively impacting ET and crop productivity \citep{kang2009climate, arora2019impact}. 
Assessing the reduction in ET, is a major challenge in agriculture. A detailed description of biophysical processes of the ET is described by the FAO-56 method \citep{allen1998crop}, using the Penman–Monteith equation, recommended for daily ET (mm/day) estimation. 
The ET under limiting condition is defined by:
\begin{equation}
    ET_a = K_s \cdot ET_x,
\end{equation}
with $K_s$ being a stress coefficient, resulting in a reduction of the ET. 
In an earlier work, the FAO described the relationship between water use and the relative yield loss \citep{doorenbos1979yield}, stating that the relative reduction in ET is related to the relative reduction in yield:
\begin{equation}\label{eq:yieldl}
    \begin{aligned}
Y_l = \left(  1-\frac{Y_a}{Y_x}\right) = K_y \left( 1- \frac{ET_a}{ET_x}\right)
\end{aligned},
\end{equation}
with $Y_a$ as the actual yield, $Y_x$ the maximum yield, or potential yield, and $K_y$ as the crop yield response factor. This factor represents the effect of a reduction in ET on the crop yield by capturing the complex relationship between crop water use and productivity. More specifically, $K_y >1$ indicates high sensitivity to water deficits with a proportionally larger yield reduction, and $K_y < 1$ indicates higher resilience to water deficits. Different studies exist, that empirically estimated $K_y$ coefficients for various crops. However, often reporting differences, making the equation difficult to solve analytically. Furthermore, $K_y$ values change over the growing period, since many crops exhibit variable susceptibility to water scarcity over the growing period. More importantly, while $ET_x$ can be more or less accurately estimated, by following \citep{allen1998crop}, estimating the $ET_a$ is is very difficult with high precision, making Eq. \ref{eq:yieldl} difficult to solve accurately in reality.  

\paragraph*{Architecture \& Evaluation}
To address this limitation, we propose to learn both $ET_a$ and $K_y$ using a Recurrent Neural Network (RNN). We constrain the network by physical information such that the reduction in ET corresponds to the observed reduction in yield. We argue that simulated $ET_x$ values are sufficiently accurate. This is supported by previous studies \citep{cai2007estimating}, allowing us to focus at $ET_a$ and $K_y$ at pixel resolution.  
We leverage an RNN, more specifically a LSTM \citep{LSTM} backbone architecture, where each hidden state is passed to a sequential layer with 128 hidden units, incorporating a linear layer, batch normalization, and dropout. Finally, two linear layers are incorporated with a single output channel each, reflecting $K_y$ and $ET_a$, respectively. Physical constraints are imposed by integrating information based on the ET and the relative reduction in crop yield, resulting in a physics-informed RNN (PI-RNN). To estimate the yield reduction at time step $t_i$ as a function of a reduction in ET, a two component loss term is employed:
\begin{equation}\label{eq:loss}
 \begin{aligned}
\mathcal{L}_{total} = \lambda_1 \mathcal{L}_{l} + \lambda_2 \mathcal{L}_{phys},
\end{aligned}
\end{equation}
\[
\text{with: } \mathcal{L}_{l} = \mathbb{E} \left[ (\hat{y_a} - y_a)^2 \right]
\] 
\[ \text{and: }
    \scriptscriptstyle{
    \mathcal{L}_{phys} = \mathbb{E} \left[ \underbrace{1_{\{ET_a < 0\}} \cdot (ET_a)^2}_{\text{lower bound penalty}} + \underbrace{1_{\{ET_a > ET_x\}} \cdot (ET_a - ET_x)^2}_{\text{upper bound penalty}} + \underbrace{1_{\{0 \leq ET_a \leq ET_x\}} \cdot (ET_a - ET_x)^2}_{\text{within bounds MSE}} \right].
    }
\]
With $1\{\cdot\}$ as the indicator function, which equals 1 if the condition inside the braces is true and 0 otherwise. The first component pushes the network to predict $K_y$ and $ET_a$ that, by using Eq. \ref{eq:yieldl} the predicted yield loss ($\hat{y_a}$) is close to the actual yield loss ($y_a$) of the ground truth data. The second component forces the network to produce $ET_a$ values bounded between [0, $ET_x$] and sufficiently close to $ET_x$, to solve Eq. \ref{eq:yieldl} with biophysical consistency. Moreover, $\lambda$ is a hyperparameter that controls the weighting of both terms. Figure \ref{fig:main_fig} (a) depicts the overall architecture of the PI-RNN.
\paragraph*{Data \& Evaluation}
As ground truth data, combine harvester yield data is used. More specifically, 54098 yield samples from 54 yield maps of cereal crops are used. This data was harvested in Switzerland between 2017 and 2021, containing georeferenced data points with information about the yield in $10m$ pixel resolution. An example yield map is depicted in Figure \ref{fig:main_fig} (b). For detailed information about the dataset and preprocessing, we refer the reader to \citep{perich2023pixel}. We define the maximum yield sample as the yield potential ($Y_x$). The reduction for each sample is expressed in relative terms as $(1 - \frac{y_a}{Y_x})$. For the simulations, we employ the FAO paper-56 \citep{allen1998crop} that simulates $ET_x$ over time. \\
For training, S2 L2A multispectral time series data is Femployed using 10 spectral bands, available from seeding to harvesting with $10m$ resolution as the model input. Additionally, weather data is incorporated, including the total precipitation and temperature, derived from \citep{hersbach2020era5}. Data modalities are early fused using the raw time series of S2 images, by aggregating weather features between S2 time steps. \\
For each experiment, a K-fold cross-validation (K=10) is performed, where results correspond to the average across folds. For quantitative evaluation, standard regression metrics are used. This includes the coefficient of determination ($R^2$), root-mean-square error (RMSE), mean absolute error (MAE). To evaluate the potential of the proposed method, we compare results against current SOTA models for crop yield prediction, including RNN (LSTM) \citep{pathak} and Transformers \citep{helber2023crop}. Note that these models cannot predict physical consistent outputs, only the final relative yield loss.

\section{Results}
\begin{figure*}[!b]
  \centering
\begin{subfigure}{.5\textwidth}
  \centering
  \includegraphics[width=\linewidth]{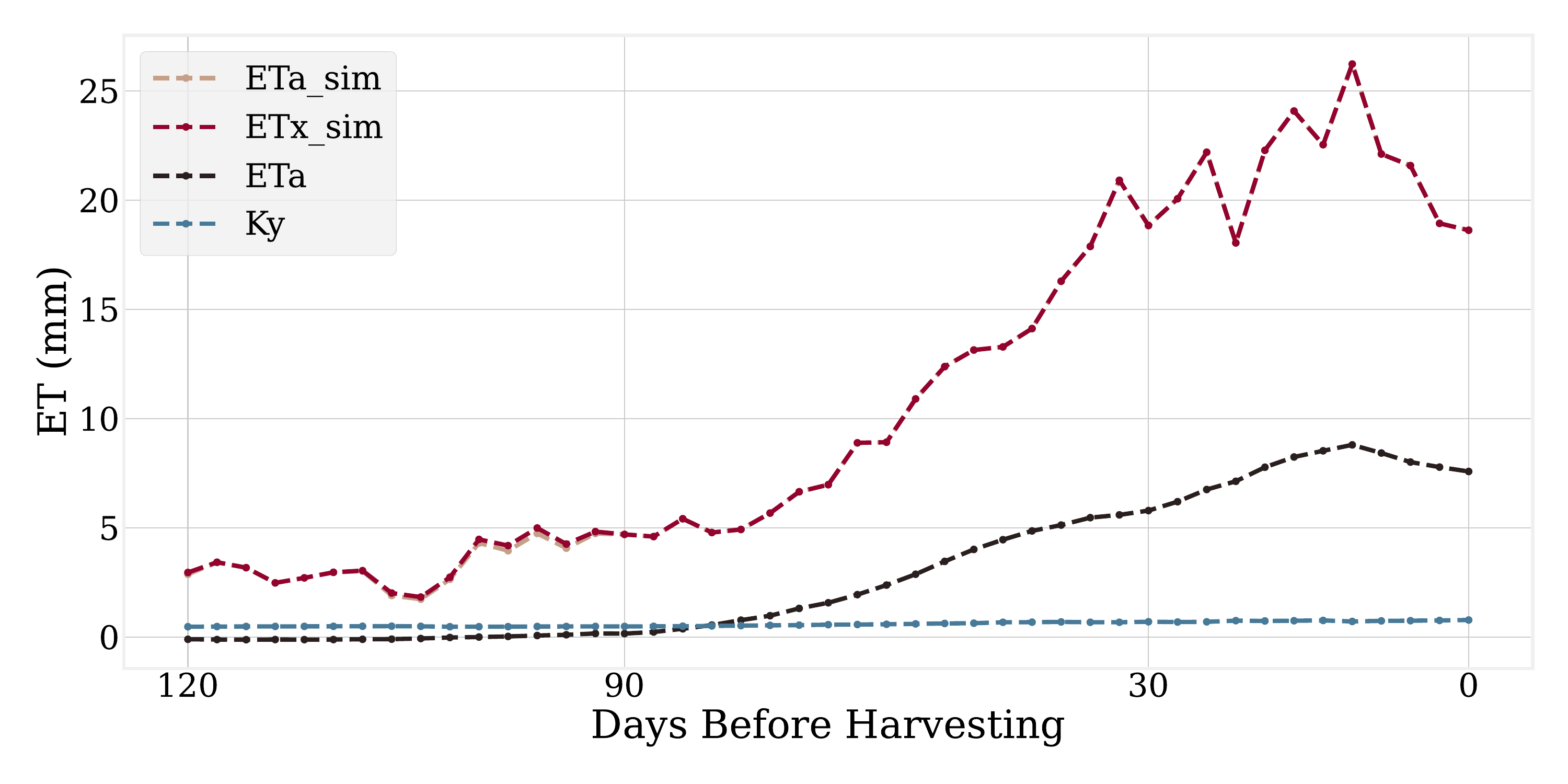}
  \caption{Predictions and simulations of crop properties.}
\end{subfigure}%
\begin{subfigure}{.5\textwidth}
  \centering
  \includegraphics[width=\linewidth]{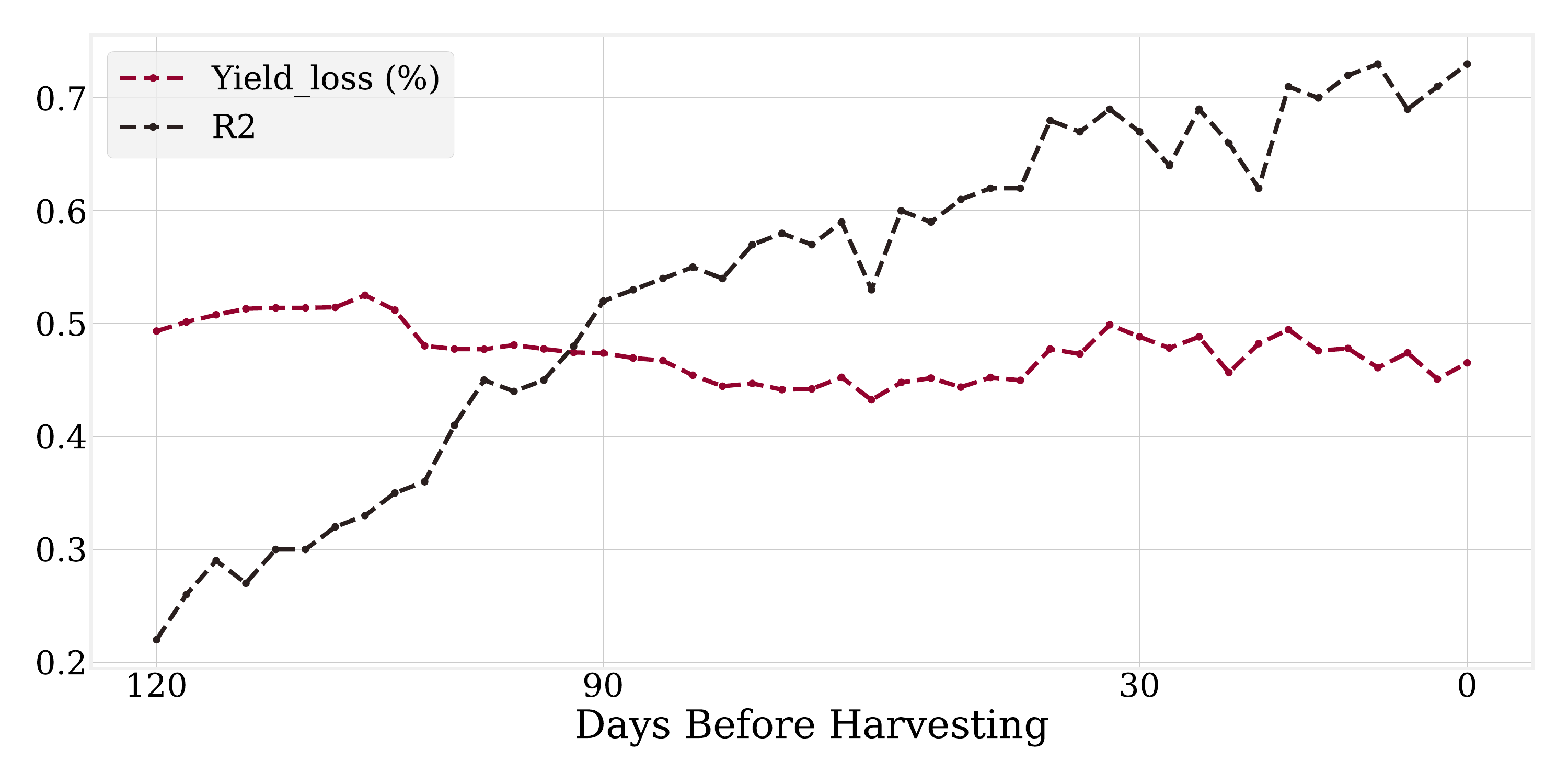}
  \caption{$R2$ scores and yield loss prediction.}
\end{subfigure}
\caption{Visualization of sequential estimations and predictions of biophysical crop properties and model performance up to 120 days before harvest.}
\label{fig:simulations}
\end{figure*}

In Table \ref{tab:model_comparision}, we evaluate the yield loss prediction performance by comparing the predicted crop yield loss to the actual yield loss and compare the results against a RNN and Transformer model. All models are trained on S2 and weather data. 
We demonstrate that our method outperforms the RNN model on various regression metrics, such as $R2$ and RMSE. In detail, we demonstrate a 2 percentage point (p.p.) improvement. Compared to the Transformer model, our method achieves similar results on all metrics. Only a minor decrease of 1 p.p, in $R2$ is observed. Additionally, we showcase the performance of the simulation model, and highlight its weak performance to predict the actual yield loss effectively. This is because the water use cannot be predicted at the pixel level, as it depends on weather data with a $30km$ spatial resolution. Additionally, the simulation models struggle to estimate actual water consumption with high accuracy (see Figure \ref{fig:simulations} (a)). Qualitatively, Figure  \ref{fig:main_fig} (b) depicts an example yield map, showcasing accurate predictions at pixel level with high spatial variability, close to the target data.
To further assess whether the inclusion of S2 data enhances the accuracy of crop yield loss estimation at the pixel level, we trained several models with different data modalities. Results are depicted in Table \ref{tab:modalities}, underlying the improved performance, when S2 data is used. In detail, a significant improvement of 21 p.p. in $R2$ is shown over the model trained solely on weather data. 
\begin{table}[!b]
    \centering
    \resizebox{.7\columnwidth}{!}{%
\begin{tabular}{llccc}
\hline
\textbf{Model} & \textbf{Modalities} & \multicolumn{1}{c}{\textbf{MAE}} & \multicolumn{1}{c}{\textbf{RMSE}} & \multicolumn{1}{c}{\textbf{R2}}  \\ \hline
RNN         & \multirow{3}{*}{Sentinel-2 + Weather} & \textbf{0.05} & 0.08          & 0.75                    \\
Transformer &                                       & \textbf{0.05} & \textbf{0.07} & \textbf{0.78}           \\
PI-RNN  (ours)    &                                       & \textbf{0.05} & \textbf{0.07} & 0.77           \\ \hline
Simulation  & Weather                               & 0.45          & 0.47          & -6.92                  \\ \hline
\end{tabular}
 
    }
    \caption{Overview table of yield loss prediction performance of different models. All models are trained using S2 and weather data. }
    \label{tab:model_comparision}
\end{table}

\begin{table}[!b]
    \centering
    \resizebox{.7\columnwidth}{!}{%
\begin{tabular}{clccc}
\hline
\multicolumn{1}{l}{\textbf{Model}} & \textbf{Modalities} & \textbf{MAE} & \textbf{RMSE} & \textbf{R2} \\ \hline
\multirow{3}{*}{PI-RNN (ours)} & Weather              & 0.08          & 0.11          & 0.46                   \\
                               & Sentinel-2           & \textbf{0.05}     & \textbf{0.07}     & 0.75         \\
                               & Sentinel-2 + Weather & \textbf{0.05} & \textbf{0.07} & \textbf{0.77} \\ \hline
\end{tabular}
 
    }
    \caption{Overview table of yield loss prediction performance of the PI-RNN (our) model, using different input modalities. }
    \label{tab:modalities}
\end{table}

Figure \ref{fig:simulations} (a) illustrates the sequential estimations of $ET_a$ and $K_y$ values up to 120 days before harvest, alongside simulated $ET_x$ and $ET_a$ values. The Figure indicates that $K_y$ values remain below 1, suggesting a higher resilience to water scarcity in this dataset. Furthermore, it is depicted that estimated $ET_a$ is consistently lower than the simulated $ET_x$, indicating yield loss due to water limitation. In Figure \ref{fig:simulations}(b), the relationship between ET and predicted yield loss is shown. The Figure demonstrates that yield loss decreases during the growing period in correlation with increased ET, as formally defined. Additionally, yield loss predictions become more accurate as the harvest approaches.  

\section{Conclusion}
Informed Neural Networks hold significant potential for crop yield modeling, offering enhanced adaptability to challenging environmental conditions. We presented a novel approach to modeling crop productivity under environmental constraints by incorporating biophysical properties and demonstrated promising experimental outcomes. To further assess the validity of the proposed method, additional experiments on larger and more diverse datasets are necessary, encompassing a wide range of yield limiting conditions. 

\bibliographystyle{unsrtnat}
\bibliography{references}  






\end{document}